\documentclass{article}

\PassOptionsToPackage{numbers, compress}{natbib}
 \usepackage[preprint]{neurips_2025}

\usepackage[utf8]{inputenc} 
\usepackage[T1]{fontenc}    
\usepackage{hyperref}       
\usepackage{url}            
\usepackage{booktabs}       
\usepackage{amsfonts}       
\usepackage{nicefrac}       
\usepackage{microtype}      
\usepackage{xcolor}         

\usepackage{longtable}
\usepackage{array}

\usepackage{graphicx}
\usepackage{subcaption}
\title{The Ouroboros of Benchmarking: Reasoning Evaluation in an Era of Saturation}

\author{%
  İbrahim Ethem Deveci \\
  Department of Cognitive Science \\
  Graduate School of Informatics \\
  Middle East Technical University \\
  Ankara, Turkey \\
  \texttt{ethem.deveci@metu.edu.tr} \\
  \And
  Duygu Ataman \\
  Department of Cognitive Science \\
  Graduate School of Informatics \\
  Middle East Technical University \\
  Ankara, Turkey \\
  \texttt{dataman@metu.edu.tr} \\
}

\begin{document}

\maketitle

\begin{abstract}
The rapid rise of Large Language Models (LLMs) and Large Reasoning Models (LRMs) has been accompanied by an equally rapid increase of benchmarks used to assess them. However, due to both improved model competence resulting from scaling and novel training advances as well as likely many of these datasets being included in pre or post training data, results become saturated, driving a continuous need for new and more challenging replacements. In this paper, we discuss whether surpassing a benchmark truly demonstrates reasoning ability or are we simply tracking numbers divorced from the capabilities we claim to measure? We present an investigation focused on three model families, OpenAI, Anthropic, and Google, and how their reasoning capabilities across different benchmarks evolve over the years. We also analyze performance trends over the years across different reasoning tasks and discuss the current situation of benchmarking and remaining challenges. By offering a comprehensive overview of benchmarks and reasoning tasks, our work aims to serve as a first reference to ground future research in reasoning evaluation and model development.
\end{abstract}

\section{Introduction}

Benchmarks have long played a central role in evaluating and comparing machine learning models \cite{benchmark}. As models scale up in size and capability, particularly Large Language Models (LLMs) and the specialized Large Reasoning Models (LRMs), many benchmarks quickly saturate, often reaching or surpassing human-level performance. Whether this saturation is driven primarily by improved model capability or dataset contamination is generally unknown. Nevertheless, this quick saturation forces the development of new and more challenging benchmarks that could be used to further compare new model families. In this paper, we investigate several key research questions: How effective are current benchmarks at measuring model capabilities, and does surpassing a benchmark reliably indicate genuine reasoning? 

To examine these questions, we select three model families, OpenAI, Anthropic, and Google, and compile performance data from official sources \cite{anthropic2024claude3,anthropic2024claude35sonnet,anthropic2025claude4,anthropic2024claude35haiku,anthropic2025claude37sonnet,anthropic2025claudeopus41,deepmind2025gemini25flashlite,google2025gemini25, deepmind2025gemini25N, geminiteam2024gemini15unlockingmultimodal, geminiteam2025geminifamilyhighlycapable, openai2024o1mini,openai2025gpt41,openai2025gpt45, openai2025gptossmodelcard,openai2025introducinggpt5,openai2025modelreleasenotes,openai2025o3o4mini, openai2024gpt4omini, openai2024helloGPT4o,openai2024learningreason}. We gather a comprehensive list of 52 benchmarks used in evaluating these models and classify them according to the types of reasoning they aim to evaluate. Analyzing performance trends over the years, we highlight where models improve, where they struggle, and what these trends reveal about the current state of benchmarking. Finally, we discuss the implications of the saturation cycle and emphasize the need for improved evaluation practices that more accurately capture model capabilities.

Our contributions are threefold: (1) we provide a curated list of reasoning benchmarks, classified by the types of reasoning they aim to assess 
(2) we analyze performance trends over the years to assess benchmarking effectiveness; (3) we examine current landscape of existing benchmarks, identifying which benchmarks have reached high performance thresholds and which seem to remain unsolved.

By situating our analysis within the broader evaluation landscape, our work collects evidence to emphasize the need for reasoning tasks that are more representative of the nature of reasoning process and target evaluation beyond downstream accuracy.

\section{Benchmark Landscape and Categorization}

\begin{figure}[ht]
    \centering
    \includegraphics[width=0.75\textwidth]{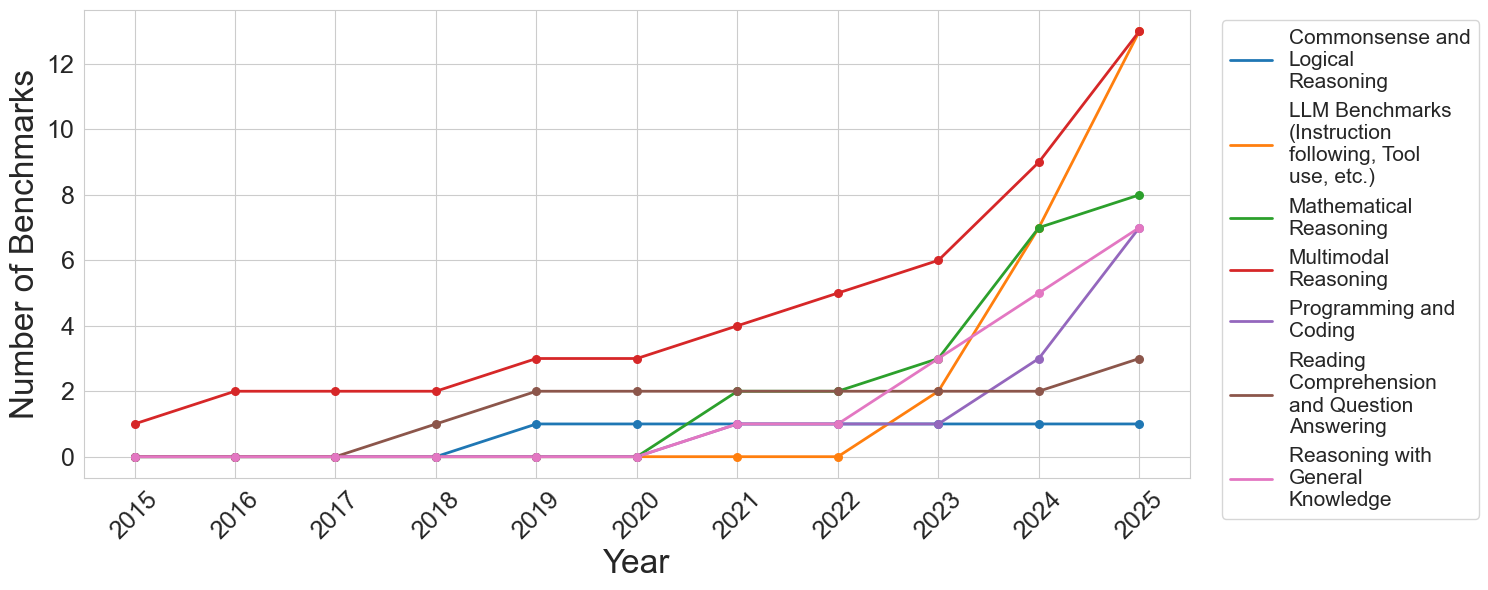} 
    \caption{Number of benchmarks in different reasoning types over time.}
    \label{fig:benchmark_landscape}
\end{figure}

In order to provide a general analysis of how the creation and adoption of reasoning benchmarks have evolved over time, we examine three model families and compile the set of benchmarks employed to evaluate them. Our aim is to provide a comprehensive overview of current benchmarking practices and to trace how the creation and adoption of benchmarks have evolved over time. The complete list of benchmarks, their assigned reasoning types, and short summaries can be found in Appendix~\ref{sec:res_benchmarks}. To facilitate analysis, we categorize benchmarks into seven reasoning types: commonsense and logical reasoning, mathematical reasoning, multimodal reasoning, programming and coding, reading comprehension and question answering, reasoning with general knowledge, and LLM-specific capabilities such as safety, tool use, and instruction following.
Figure \ref{fig:benchmark_landscape} illustrates a marked increase in benchmark adoption for multimodal reasoning, mathematical reasoning, programming, reasoning with general knowledge, and LLM-specific benchmarks after 2023. In contrast, no new benchmarks in reading comprehension or commonsense reasoning were adopted by these model families during this period. While the literature contains several other benchmarks in these areas \cite{commonsense1, commonsense2, commonsense3, commonsense4, commonsense5, reading1, reading2}, our analysis shows they have not been utilized by any of the prominent model families. This likely reflects the evolving understanding of what constitutes reasoning in computational models, in accordance with their current capabilities and what the community deems important to evaluate. Since most models now have direct commercial applications, their performance in more applicable domains, such as coding and tool-use benchmarks, may also motivate the evaluation in certain categories of reasoning tasks.


\section{Performance Trends Across Models}

Across all three model families there is a consistent effort to develop newer models or architectural improvements to achieve higher benchmark performance. However, comparing performance across families is challenging, as each family often employs different benchmarks, and even within a single family, benchmarks used can vary between model iterations. This variation appears to stem from two main factors: first, certain benchmarks reach saturation due to high performance; second, benchmark updates or more challenging subsets are introduced, such as the transition from MATH to MATH-500 \cite{MATH-500}.

We observe a recurring pattern: once a model family achieves a high performance on a particular benchmark, subsequent models tend to use that benchmark less frequently or may discontinue its use entirely. This reflects both practical and conceptual considerations: benchmarks that no longer discriminate between models provide limited evaluative value, and benchmark selection increasingly reflects the evolving understanding of which reasoning tasks remain challenging for current architectures.

Interestingly, performance trends reveal consistent directional correlations across benchmarks within the same reasoning type. For example, when a model demonstrates improved performance on a benchmark, it generally shows corresponding improvements on other benchmarks of the same type, while lower performance on one benchmark tends to coincide with lower performance on others. Nevertheless, the extent of performance differs across benchmarks, potentially due to variations in problem complexity and the scaling limitations evident in smaller models, as seen within the OpenAI family. This pattern suggests that benchmarks within a reasoning type often capture overlapping aspects of reasoning, so that advances in a models' capabilities tend to propagate across related tasks. At the same time, variations in the magnitude of performance gains provide insight into the relative difficulty of different benchmarks within the same reasoning type. Detailed plots illustrating performance changes within model families for different reasoning types are provided in Appendix~\ref{sec:model_performance}.

Finally, we note that newer models generally achieve higher performance on previously low-scoring benchmarks. However, the limited overlap of common benchmarks across model families complicates cross-family comparisons. This raises a critical question: if benchmarks are intended to evaluate and compare model capabilities, why are they not consistently adopted or reported across families? If benchmarks are intended to provide a shared measure of capability, their fragmented and selective use undermines that goal and exemplifies the need for more standardized, representative, and domain-informed evaluation frameworks.

\section{Performance of Models within Benchmarks}

We collect all reported model performances across benchmarks and analyze saturation by defining it as whether a model has achieved at least 80\% accuracy on the given benchmark. Out of the full set of benchmarks, we find that 27 benchmarks surpass this threshold in at least one model family, while 25 benchmarks never reach it. The majority of “solved” benchmarks belong to commonsense and logical reasoning, mathematical reasoning, reasoning with general knowledge, and reading comprehension and question answering. By contrast, benchmarks targeting LLM-specific capabilities and programming and coding remain comparatively difficult, with few instances of performance above 80\%.

We then examine the release years of benchmarks that never surpass the 80\% threshold. The distribution is striking: 60\% of unsolved benchmarks were introduced in 2025, 32\% in 2024, and only two pre-2023 benchmarks remain unsolved, which are ActivityNet \cite{ACTIVITYNET} and EgoSchema \cite{EGOSCHEMA}, both multimodal reasoning benchmarks. This distribution suggests a clear trend. Nearly all benchmarks released prior to 2025 have already been surpassed by at least one model family, indicating rapid saturation. By contrast, the benchmarks still below the threshold overwhelmingly correspond to the most recently introduced evaluation tasks.

\begin{figure}[ht]
    \centering
    \begin{subfigure}[t]{0.48\textwidth}
        \centering
        \includegraphics[width=\textwidth]{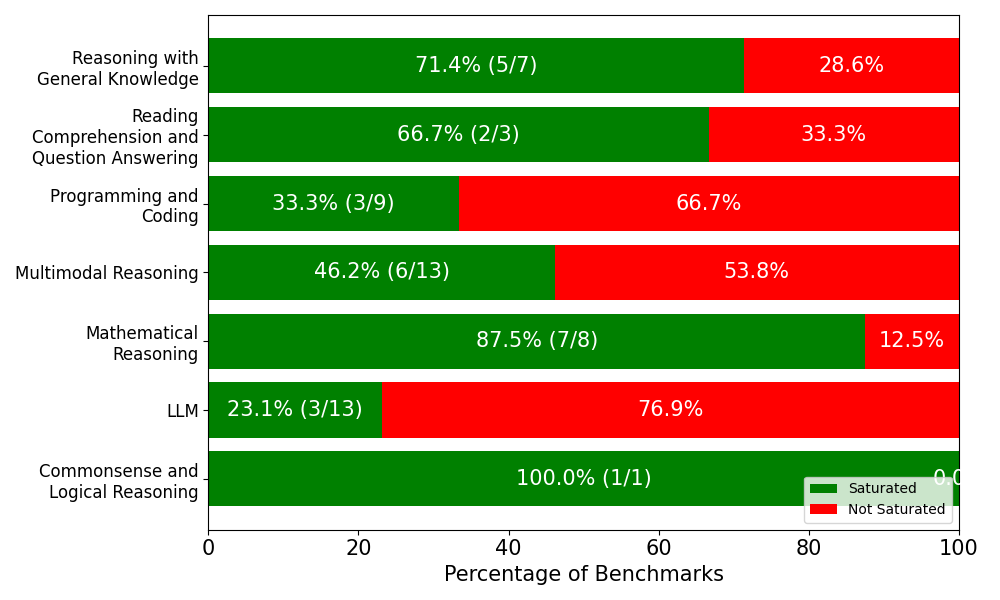}
        \caption{Distribution of benchmarks that models surpassed 80\% threshold and those not yet surpassed, grouped by reasoning type.}
        \label{fig:benchmark_saturation}
    \end{subfigure}
    \hfill
    \begin{subfigure}[t]{0.50\textwidth}
        \centering
        \includegraphics[width=\textwidth]{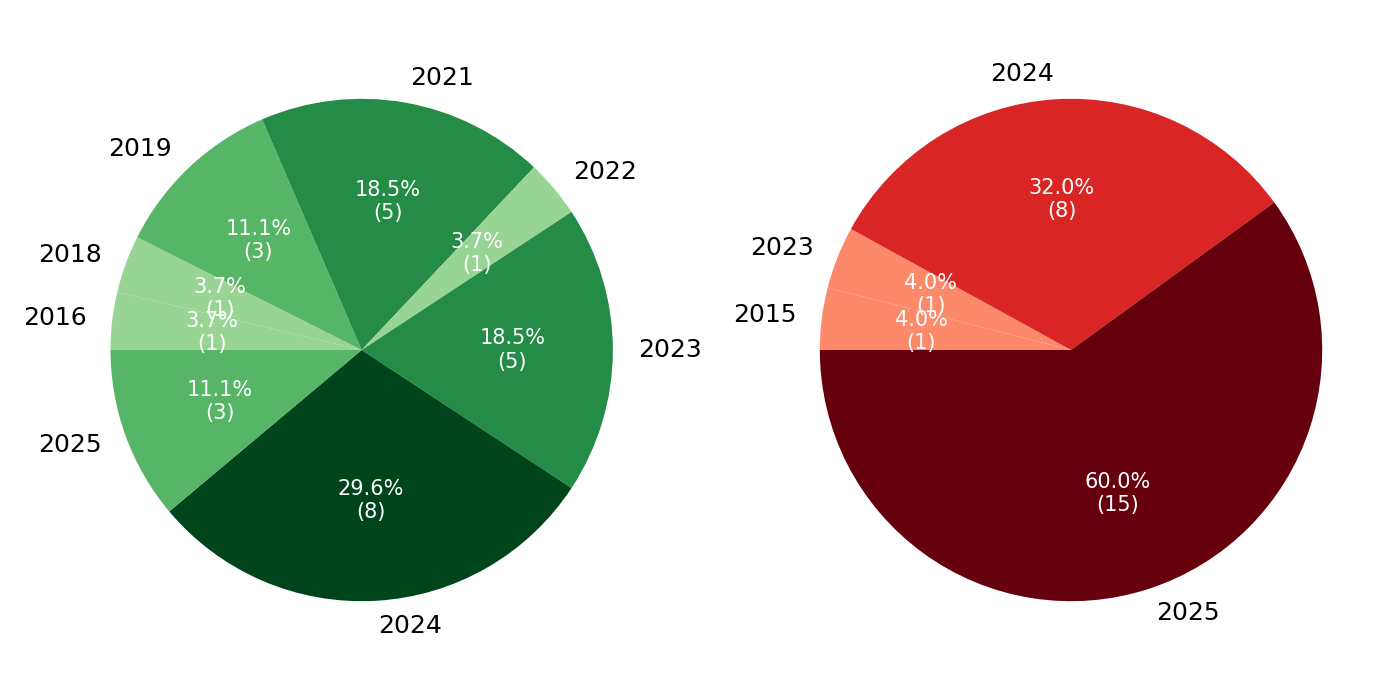}
        \caption{Release years of benchmarks relative to the 80\% threshold: left pie shows surpassed benchmarks, right pie shows unsolved benchmarks.}
        \label{fig:benchmark_saturation_year}
    \end{subfigure}
    \caption{Benchmark saturation dynamics.}
    \label{fig:benchmark_saturation_combined}
\end{figure}

This temporal pattern highlights the central dynamic of the saturation cycle: older benchmarks are rapidly mastered and lose discriminative power, while newly introduced benchmarks become the standards for demonstrating progress. Nearly all unsolved benchmarks are recent, highlighting both the accelerating pace of benchmark creation and the difficulty of maintaining evaluations that remain challenging over time. Yet this difficulty seems only temporary. It is highly plausible that within one or two years many of these currently unsolved benchmarks will also be surpassed, at which point model families will shift to alternative or newly designed evaluations to preserve differentiation. Crucially, this pattern reflects the fact that performance gains are often specific to individual benchmarks rather than to the broader reasoning type they are intended to assess. As the analyses indicate, while models often perform consistently and even strongly on benchmarks within a domain, the introduction of a more challenging, novel benchmark frequently leads to a drop in performance. This pattern may arise from the increased difficulty of the new benchmark, or from contamination that inflated performance on earlier benchmarks without truly reflecting generalizable reasoning ability. This situation raises the question of whether what appears as “reasoning ability” is often tied more to benchmark design and prior exposure than to robust mastery of the reasoning type itself. This saturation cycle casts doubt on the long-term evaluation value of benchmarks.

\section{Discussion: Limitations of Current Benchmarking}

Our analysis of three model families demonstrates that benchmark performance has generally increased over time, with newer models achieving higher scores across most reasoning types and benchmarks. However, given that many benchmarks have already been surpassed with high accuracy, we would like to highlight a question originally posed in \cite{commonsense3} regarding commonsense reasoning, reframed here for reasoning in general: \textit{Have neural language models successfully acquired reasoning, or are we overestimating the true capabilities of machine reasoning?}
Several studies in the literature show that these models still perform poorly when required to generalize to longer contexts or handle tasks requiring inductive and compositional reasoning \cite{ind1, ind2, ind3, ind4, ind5, ind6}. This discrepancy suggests a limitation of current benchmarking practices: improvements in benchmark scores do not necessarily reflect generalizable reasoning ability. 

We believe this discrepancy can be reduced by developing more sophisticated, task-specific evaluation metrics that capture intermediate reasoning steps or different modes of error. Additionally, formalizing reasoning for different task types can support these efforts, enabling more structured analyses and clearer assessment of models’ reasoning abilities. Such a formalization enables structured representations of diverse reasoning types and their interrelationships \cite{formal1, formal2, formal3}, and facilitates the design of layered, targeted evaluation procedures that assess specific reasoning capabilities rather than merely reporting overall accuracy. Furthermore, formal reasoning frameworks can support the development of algorithms that deliver structured feedback to models, guiding the refinement of their reasoning abilities. By integrating formalized reasoning with task-specific evaluations, benchmarking can be conducted in a more targeted and informative manner.

\section{Limitations}

The analysis in our study focuses on 52 benchmarks used by the three model families. Other model families and reasoning-focused models are not fully explored because including them, along with more than two hundred benchmarks identified from other model families and several studies evaluating different types of reasoning in large models, would create a combinatorial explosion of comparisons. This restriction was necessary to maintain the scope of our work on a qualitative evaluation of benchmark design and adoption rather than an exhaustive quantitative analysis of all models and benchmarks. A comprehensive comparison across a wider range of models and benchmarks is left for future work.

\section{Conclusion}

In this work, we analyze 52 benchmarks across three model families, covering multiple reasoning types. Our study reveals the rapid saturation of older benchmarks, selective adoption of new ones, and temporal dynamics that govern the utility of benchmarks in evaluating model performance. While model performance generally improves over time and correlations within reasoning types indicate overlapping evaluation properties, the introduction of more challenging benchmarks generally resets performance, suggesting that apparent reasoning ability is influenced more by extrinsic factors than by mastering the reasoning itself, as supported by other studies. This saturation cycle highlights the limitations of current practices: benchmarks provide only a partial view of model reasoning. Meaningful progress requires formalized reasoning tasks, layered evaluation procedures, and task-specific metrics that go beyond accuracy scores.

\bibliographystyle{unsrt}
\bibliography{references}


\appendix

\clearpage
\section{Reasoning Benchmarks}
\label{sec:res_benchmarks}
\renewcommand{\arraystretch}{1.2}
\setlength{\tabcolsep}{6pt}

\begin{longtable}{|>{\raggedright\arraybackslash}p{2 cm}|
                        >{\raggedright\arraybackslash}p{2.2cm}|
                        >{\centering\arraybackslash}p{1.2cm}|
                        >{\raggedright\arraybackslash}p{6cm}|}
\caption{Taxonomy of benchmarks used in this study.} \label{tab:reasoning-benchmarks} \\

\hline
\textbf{Benchmark} & \textbf{Reasoning Type} & \textbf{Year} & \textbf{Explanation} \\
\hline
\endfirsthead

\hline
\textbf{Benchmark} & \textbf{Reasoning Type} & \textbf{Year} & \textbf{Explanation} \\
\hline
\endhead

HellaSwag \cite{hellaswag} & Commonsense and Logical Reasoning & 2019 & Multiple-choice task: choose the most plausible sentence continuation. \\ \hline
MMLU \cite{MMLU} & Reasoning with General Knowledge & 2021 & Multiple-choice task: answer questions across 57 domains to test knowledge and problem-solving. \\ \hline
Big-Bench-Hard \cite{BIG-BENCH-HARD} & Reasoning with General Knowledge & 2023 & Open-generation task: solve difficult BIG-Bench problems testing multi-step reasoning and problem-solving. \\ \hline
MMMLU \cite{MMMLU} & Reasoning with General Knowledge & 2024 & Multiple-choice task: answer 57 domain questions translated into 14 languages to test multilingual knowledge and problem-solving. \\ \hline
Humanity's Last Exam \cite{HUMANITY'SLE} & Reasoning with General Knowledge & 2025 & Multi-modal task: answer closed-ended questions across many subjects to test verifiable knowledge. \\ \hline
Global MMLU (Lite) \cite{GLOBALMMLU(LITE)} & Reasoning with General Knowledge & 2025 & Multiple-choice task: answer 42-language questions with culturally sensitive labeling to test equitable multilingual knowledge. \\ \hline
GPQA Diamond \cite{GPQADIAMOND} & Reasoning with General Knowledge & 2023 & Multiple-choice task: answer 448 expert-level science questions in biology, physics, and chemistry that are Google-proof and highly challenging. \\ \hline
MMLU Pro \cite{MMLUPRO} & Reasoning with General Knowledge & 2024 & Multiple-choice task: extended from MMLU, answer more challenging reasoning questions with 10 options across diverse domains. \\ \hline
ARC (AI2 Reasoning Challenge) \cite{ARCAI2} & Reading Comprehension and Question Answering & 2018 & Multiple-choice task: answer grade-school science questions requiring advanced knowledge and reasoning beyond simple retrieval. \\ \hline
ECLeKTic \cite{ECLECTIC} & Reading Comprehension and Question Answering & 2025 & Closed-book QA task: answer 12-language questions to test cross-lingual knowledge transfer. \\ \hline
DROP \cite{DROP} & Reading Comprehension and Question Answering & 2019 & Open-ended QA task: answer 96k English questions requiring discrete reasoning over paragraph content. \\ \hline
GSM8K \cite{GSM8K} & Mathematical Reasoning & 2021 & Open-ended QA task: solve grade-school problems requiring multi-step mathematical reasoning. \\ \hline
MATH \cite{MATH-500} & Mathematical Reasoning & 2021 & Open-ended QA: solve 12,500 challenging competition problems with step-by-step solutions to test advanced mathematical reasoning. \\ \hline
MATH 500 \cite{MATH-500} & Mathematical Reasoning & 2024 & Open-ended QA: Challenging subset of MATH benchmark. \\ \hline
MGSM \cite{MGSM} & Mathematical Reasoning & 2023 & Open-ended QA: solve 250 GSM8K problems translated into 10 languages. \\ \hline
MathVista \cite{MATHVISTA} & Mathematical Reasoning & 2024 & Open-ended multimodal QA: solve 6,141 math problems requiring visual and compositional reasoning. \\ \hline
AIME 2024 & Mathematical Reasoning & 2024 & Open-ended QA: solve challenging competition-level mathematics problems. \\ \hline
AIME 2025 & Mathematical Reasoning & 2025 & Open-ended QA: solve challenging competition-level mathematics problems. \\ \hline
FrontierMath \cite{FRONTIERMATH} & Mathematical Reasoning & 2024 & Open-ended QA: tests advanced mathematical reasoning across diverse and expert-level domains, requiring multi-step problem solving and deep mathematical knowledge. \\ \hline
MMMU \cite{MMMU} & Multimodal Reasoning & 2024 & Question answering task: multimodal multiple-choice and open-ended questions across 30 subjects requiring advanced reasoning and domain-specific knowledge. \\ \hline
AI2D \cite{AI2D} & Multimodal Reasoning & 2016 & Open-ended QA: multimodal questions with 5,000 diagrams and 15,000 Q\&A pairs requiring diagram structure understanding and reasoning. \\ \hline
ChartQA \cite{CHARTQA} & Multimodal Reasoning & 2022 & Open-ended QA: multimodal questions with 32.7K chart-based problems requiring visual and logical reasoning. \\ \hline
EgoSchema \cite{EGOSCHEMA} & Multimodal Reasoning & 2023 & Multiple-choice QA: multimodal questions with 5,000 long-form video clips requiring understanding of human activity and temporal reasoning. \\ \hline
DocVQA \cite{DOCVQA} & Multimodal Reasoning & 2021 & Open-ended QA: multimodal questions with 50,000 document images requiring reading and interpreting document layout and structure. \\ \hline
TextVQA \cite{TEXTVQA} & Multimodal Reasoning & 2019 & Open-ended QA: multimodal questions with 45,336 images requiring reading and reasoning about embedded text. \\ \hline
VideoMMMU \cite{VIDEOMMMU} & Multimodal Reasoning & 2025 & Open-ended QA: multimodal questions with 300 expert-level videos and 900 Q\&A pairs assessing knowledge acquisition through perception, comprehension, and adaptation. \\ \hline
Vibe-Eval \cite{VIBEEVAL} & Multimodal Reasoning & 2024 & Open-ended QA: multimodal questions, testing visual understanding and multimodal chat capabilities. \\ \hline
ZeroBench \cite{ZEROBENCH} & Multimodal Reasoning & 2025 & Open-ended QA: multimodal questions with 434 visual reasoning problems designed to be impossible for current LMMs. \\ \hline
CharXiv \cite{CHARXIV} & Multimodal Reasoning & 2024 & Open-ended QA: multimodal questions with 2,323 charts requiring descriptive analysis and complex reasoning. \\ \hline
MMMU Pro \cite{MMMUPRO} & Multimodal Reasoning & 2025 & QA task: multimodal multiple-choice and open-ended questions, extended from MMMU, testing integrated visual and textual reasoning. \\ \hline
ActivityNet \cite{ACTIVITYNET} & Multimodal Reasoning & 2015 & Multiple-choice and open-ended QA: evaluates recognition and understanding of complex human activities in untrimmed videos, testing visual perception and temporal reasoning. \\ \hline
ERQA \cite{ERQA} & Multimodal Reasoning & 2025 & Multiple-choice QA: evaluates embodied reasoning and spatial understanding in real-world scenarios, requiring models to integrate text and visual inputs to select the correct answer. \\ \hline
SWE-bench Verified \cite{SWEBENCH} & Programming and Coding & 2024 & Open-ended QA: answer 2,294 software engineering problems requiring multi-file code edits and complex reasoning. \\ \hline
Terminal-bench \cite{TERMINALBENCH} & Programming and Coding & 2025 & Open-ended QA: answer complex tasks in terminal environments using text-based commands and reasoning. \\ \hline
HumanEval \cite{HUMANEVAL} & Programming and Coding & 2021 & Open-ended QA: answer Python programming problems from docstrings requiring functional code synthesis. \\ \hline
LiveCode Bench \cite{LIVECODEBENCH} & Programming and Coding & 2025 & Open-ended QA: answer 600+ coding problems from contests, testing generation, self-repair, execution, and test prediction. \\ \hline
Aider Polygot \cite{AIDERPOLYGOT} & Programming and Coding & 2024 & Open-ended QA: answer 225 difficult coding problems in C++, Go, Java, JavaScript, Python, and Rust. \\ \hline
SWE-Lancer \cite{SWELANCER} & Programming and Coding & 2025 & Open-ended QA: answer 1,400 freelance software engineering tasks, including implementation and managerial decisions, with real-world evaluation. \\ \hline
SWE-Lancer Diamond \cite{SWELANCER} & Programming and Coding & 2025 & Open-ended QA: answer tasks from the public SWE-Lancer Diamond split, including implementation and managerial software engineering problems. \\ \hline
TAU-bench \cite{TAUBENCH} & Tool Use -- LLM & 2024 & Open-ended QA: tests reasoning, consistency, and rule-following in dynamic, tool-assisted human-agent interactions. \\ \hline
TAU2-bench \cite{TAU2BENCH} & Tool Use -- LLM & 2025 & Open-ended QA: tests multi-turn reasoning, coordination, and communication in dual-control environments where both agent and user act with tools. \\ \hline
COLLIE \cite{COLLIE} & Constrained Text Generation -- LLM & 2023 & Open-ended QA: answer 2,080 prompts requiring constrained text generation with compositional, grammar-based, and reasoning challenges. \\ \hline
SimpleQA \cite{SIMPLEQA} & Factuality -- LLM & 2024 & Factual QA benchmark designed to test factual accuracy and knowledge calibration. \\ \hline
FACTS Grounding \cite{FACTS} & Factuality -- LLM & 2024 & Open-ended QA: answer questions requiring LLMs to generate factually accurate and well-grounded responses from provided source material. \\ \hline
BrowseComp \cite{BROWSECOMP} & Factuality -- LLM & 2025 & Open-ended QA: answer 1,266 questions by persistently navigating the internet to find hard-to-locate information. \\ \hline
ComplexFunc Bench \cite{COMPLEXFUNCBENCH} & Tool Use -- LLM & 2025 & Open-ended QA: answer complex function-calling tasks in five real-world scenarios requiring multi-step reasoning, parameter management, and long-context handling. \\ \hline
IFEval \cite{IFEVAL} & Instruction Following -- LLM & 2023 & Open-ended QA: answer 500 prompts requiring LLMs to follow verifiable natural language instructions. \\ \hline
Multi-IF \cite{MULTIIF} & Instruction Following -- LLM & 2024 & Open-ended QA: answer 4,501 multilingual multi-turn prompts requiring accurate instruction-following across languages and conversation turns. \\ \hline
LOFT \cite{LOFT} & Long-Context -- LLM & 2024 & Open-ended QA: answer real-world tasks requiring reasoning and in-context retrieval over millions of tokens. \\ \hline
Graphwalks \cite{openai2025gpt41} & Long-Context -- LLM & 2025 & Open-ended QA: perform multi-hop reasoning across a graph of millions of tokens to answer questions requiring breadth-first traversal. \\ \hline
Multi Challenge \cite{MULTICHALLENGE} & Multi-turn Conversation -- LLM & 2025 & Open-ended QA: answer multi-turn conversation prompts requiring instruction-following, context management, and in-context reasoning. \\ \hline
HealthBench \cite{HEALTHBENCH} & Safety -- LLM & 2025 & Open-ended QA: evaluates LLMs on multi-turn healthcare conversations, requiring factual reasoning, safety awareness, and context-sensitive decision-making across diverse medical contexts. \\ \hline

\end{longtable}

\clearpage
\section{Performance of Models}
\label{sec:model_performance}

\begin{figure}[htp]
    \centering
    
    \begin{subfigure}{0.40\textwidth}
        \centering
        \includegraphics[width=\linewidth]{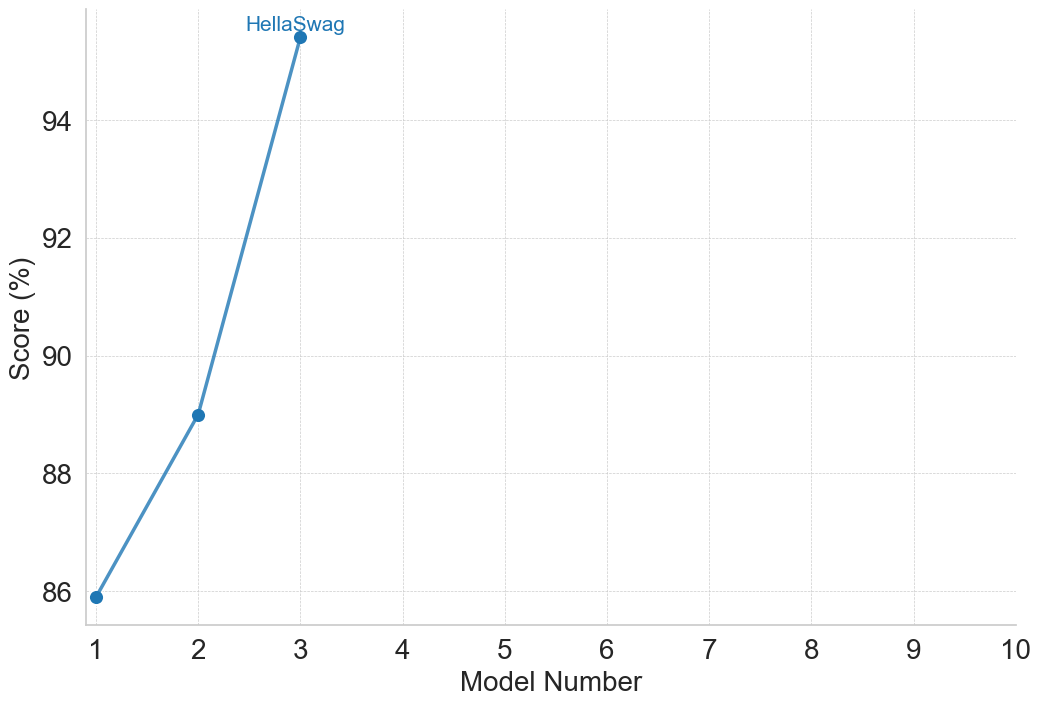}
        \caption{Commonsense and Logical Reasoning}
        \label{fig:resA}
    \end{subfigure}
    \hfill
    \begin{subfigure}{0.40\textwidth}
        \centering
        \includegraphics[width=\linewidth]{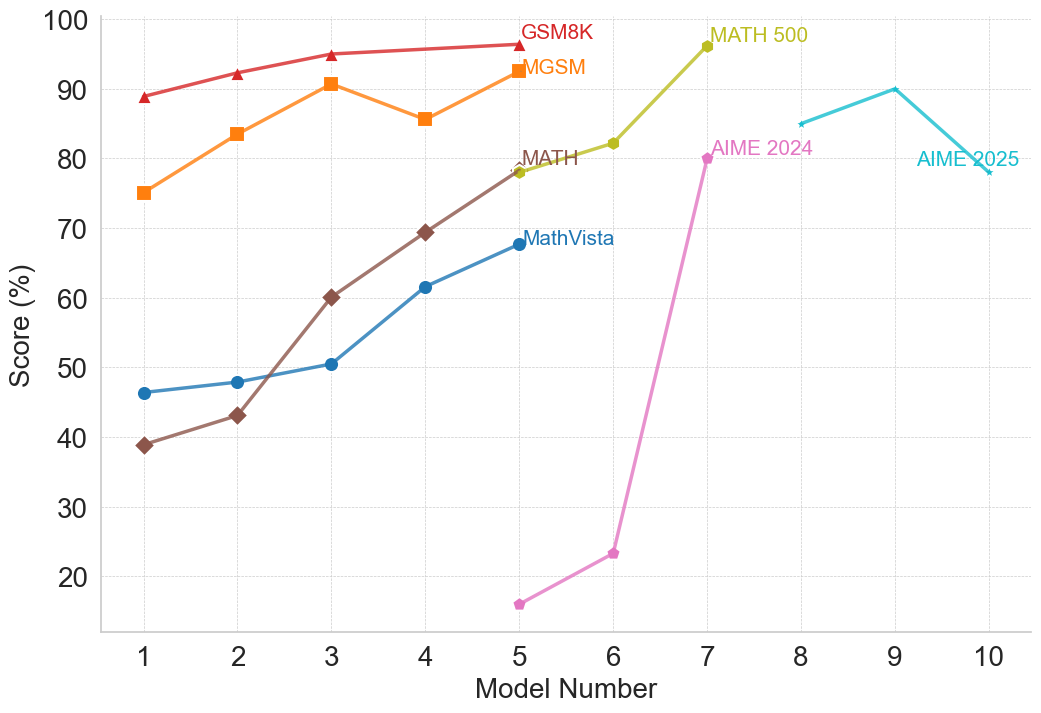}
        \caption{Mathematical Reasoning}
        \label{fig:resB}
    \end{subfigure}
    
    \vspace{0.5em}
    \begin{subfigure}{0.40\textwidth}
        \centering
        \includegraphics[width=\linewidth]{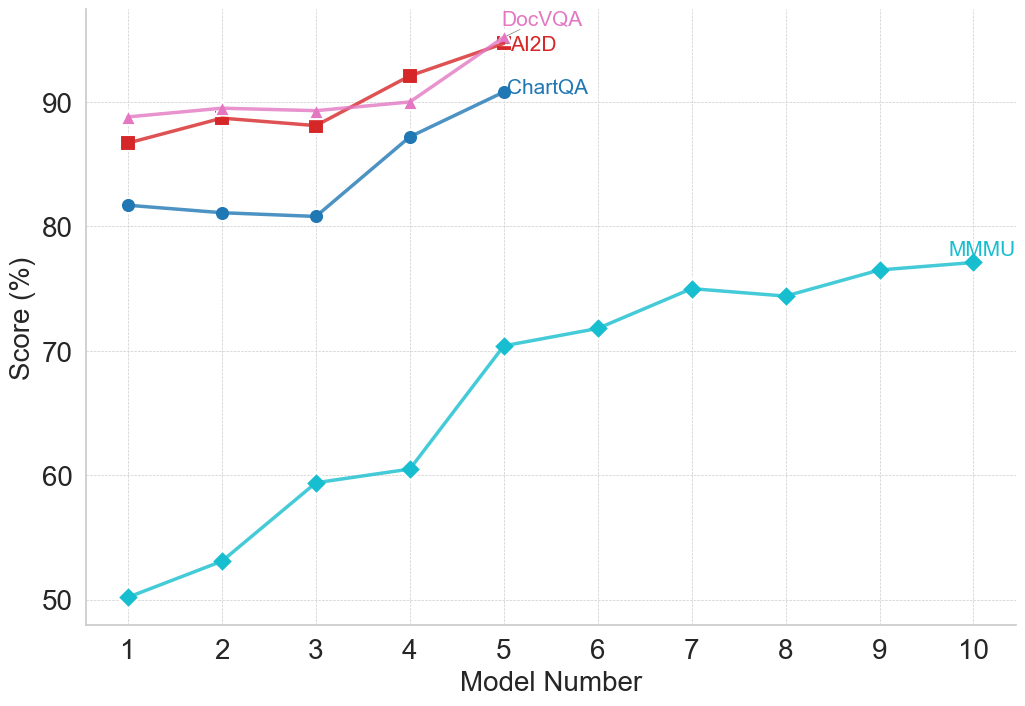}
        \caption{Multimodal Reasoning}
        \label{fig:resC}
    \end{subfigure}
    \hfill
    \begin{subfigure}{0.40\textwidth}
        \centering
        \includegraphics[width=\linewidth]{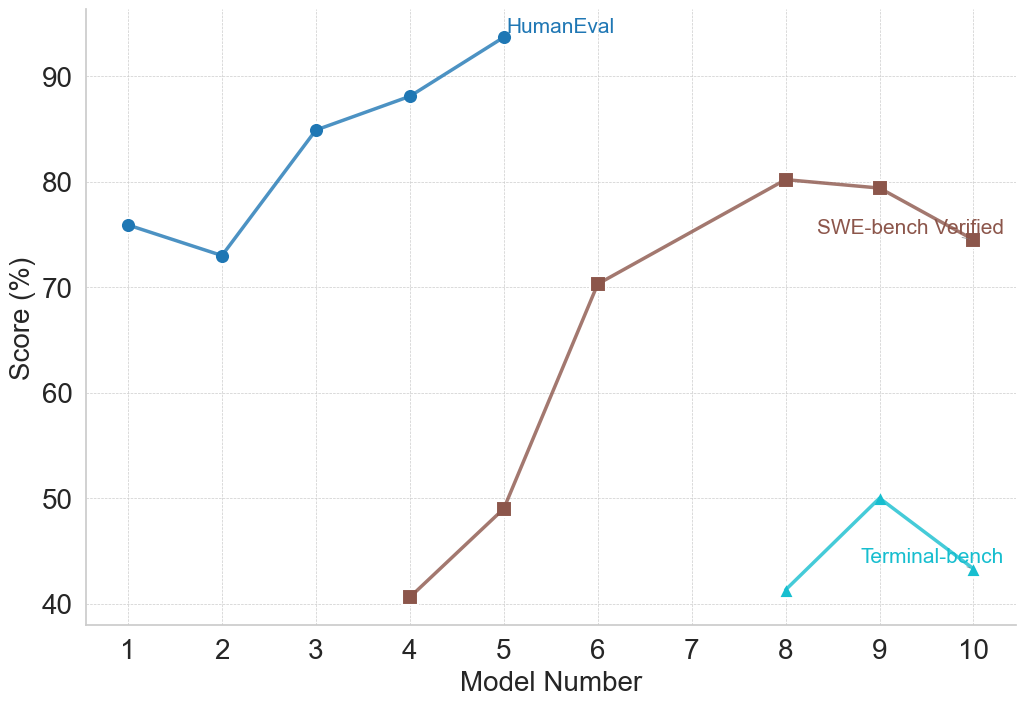}
        \caption{Programming and Coding}
        \label{fig:resD}
    \end{subfigure}
    
    \vspace{0.5em}
    \begin{subfigure}{0.40\textwidth}
        \centering
        \includegraphics[width=\linewidth]{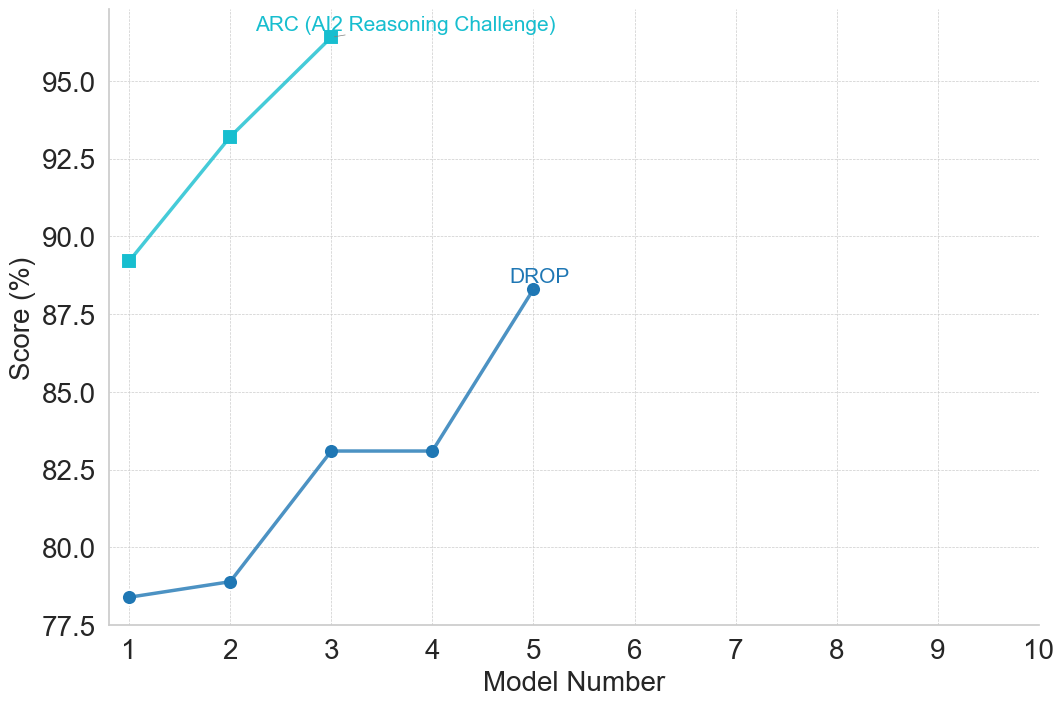}
        \caption{Reading Comprehension and QA}
        \label{fig:resE}
    \end{subfigure}
    \hfill
    \begin{subfigure}{0.40\textwidth}
        \centering
        \includegraphics[width=\linewidth]{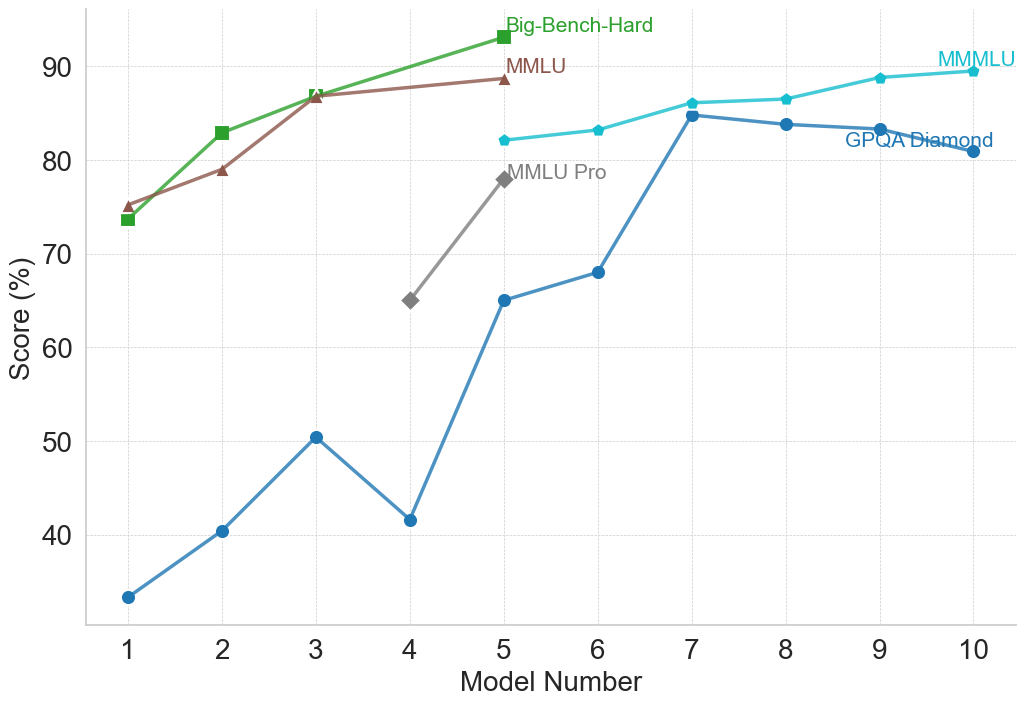}
        \caption{Reasoning with General Knowledge}
        \label{fig:resF}
    \end{subfigure}
    
    \vspace{0.5em}
    \begin{subfigure}{0.40\textwidth}
        \centering
        \includegraphics[width=\linewidth]{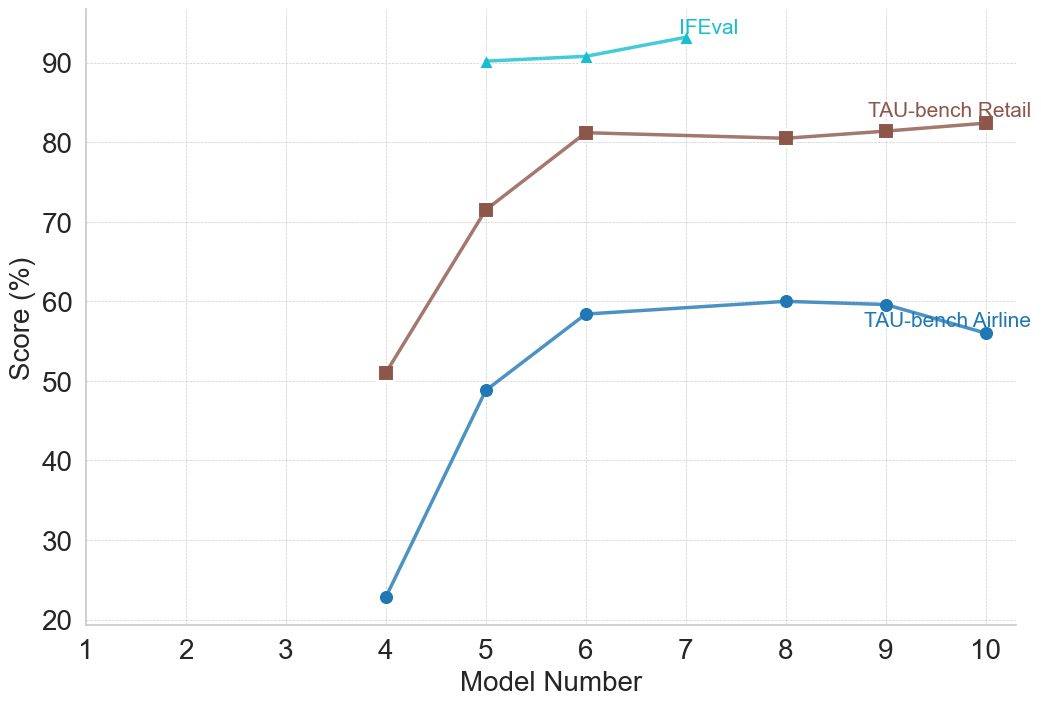}
        \caption{LLM Benchmarks}
        \label{fig:resG}
    \end{subfigure}

\caption{
Performance of the Claude family on reasoning benchmarks by category. 
Model numbers and corresponding names are as follows: 
1 -- Claude 3 Haiku; 
2 -- Claude 3 Sonnet; 
3 -- Claude 3 Opus; 
4 -- Claude 3.5 Haiku; 
5 -- Claude 3.5 Sonnet; 
6 -- Claude 3.7 Sonnet; 
7 -- Claude 3.7 Sonnet (64K Extended Thinking); 
8 -- Claude Sonnet 4; 
9 -- Claude Opus 4; 
10 -- Claude Opus 4.1.
}
    \label{fig:claude-grid}
\end{figure}

\begin{figure}[htp]
    \centering
    
    \begin{subfigure}{0.45\textwidth}
        \centering
        \includegraphics[width=\linewidth]{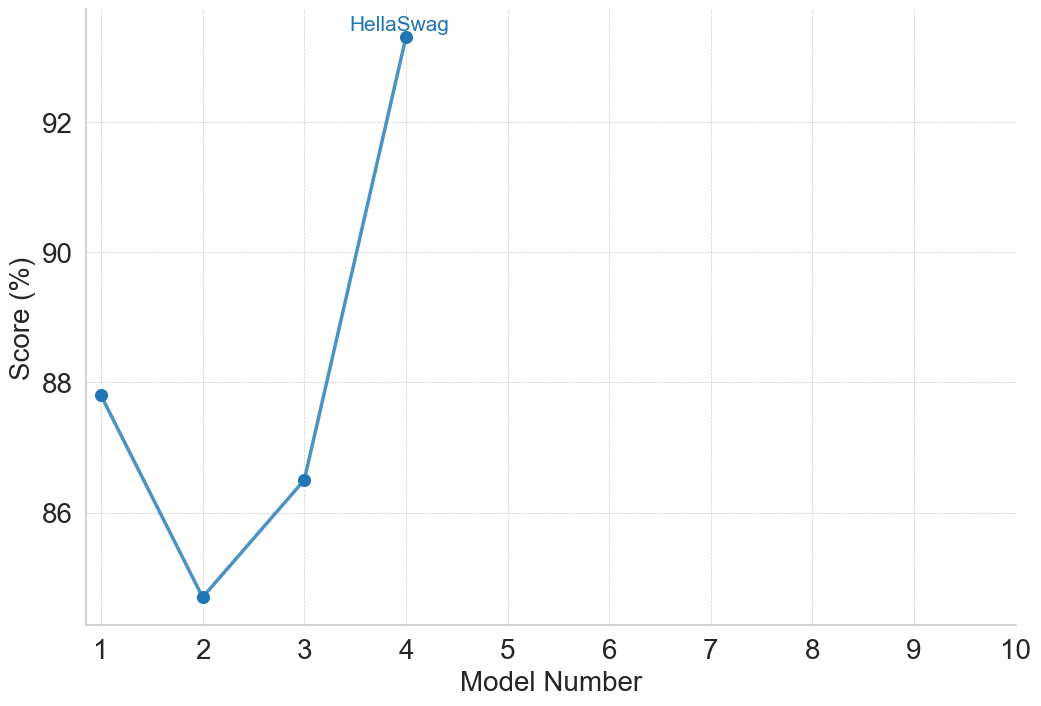}
        \caption{Commonsense and Logical Reasoning}
        \label{fig:resH}
    \end{subfigure}
    \hfill
    \begin{subfigure}{0.45\textwidth}
        \centering
        \includegraphics[width=\linewidth]{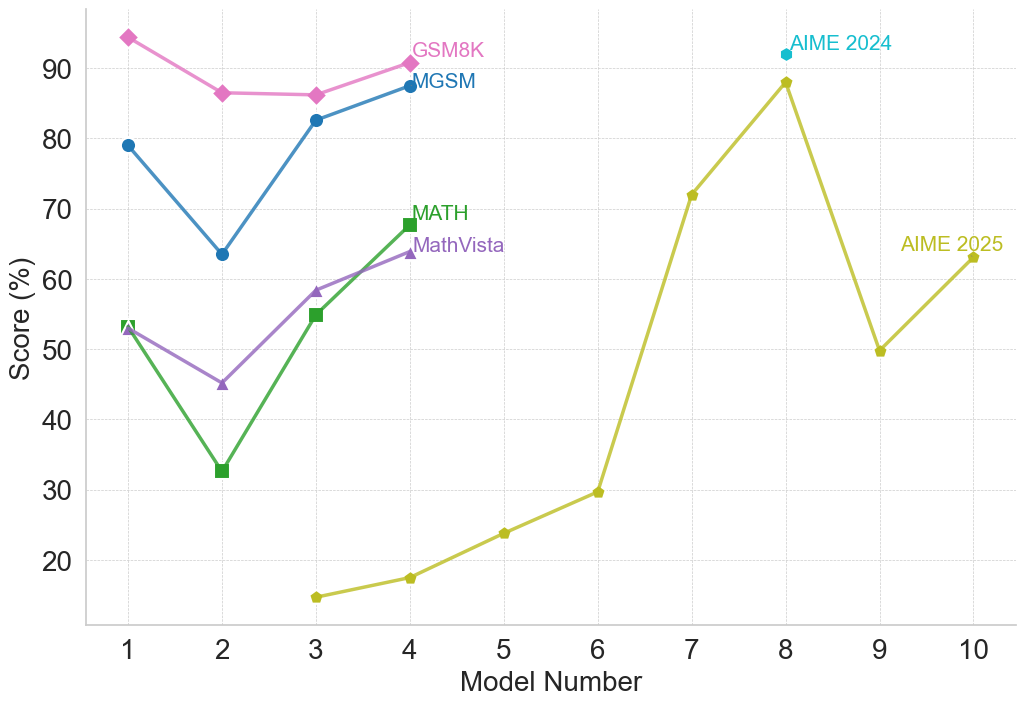}
        \caption{Mathematical Reasoning}
        \label{fig:resI}
    \end{subfigure}

    \vspace{0.5em}
    \begin{subfigure}{0.45\textwidth}
        \centering
        \includegraphics[width=\linewidth]{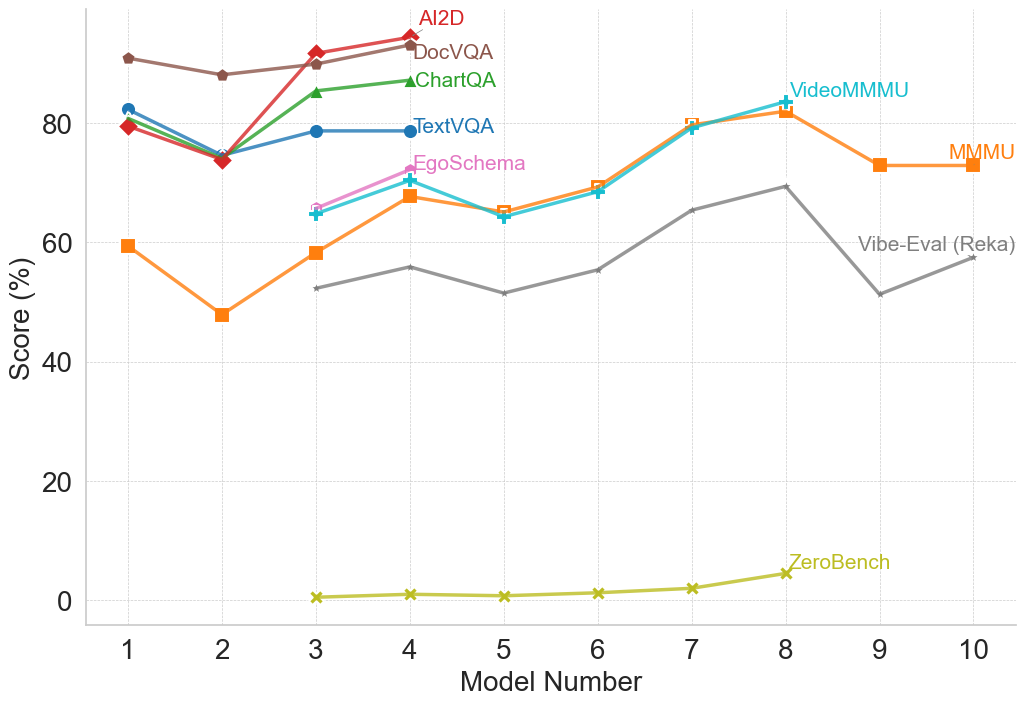}
        \caption{Multimodal Reasoning}
        \label{fig:resJ}
    \end{subfigure}
    \hfill
    \begin{subfigure}{0.45\textwidth}
        \centering
        \includegraphics[width=\linewidth]{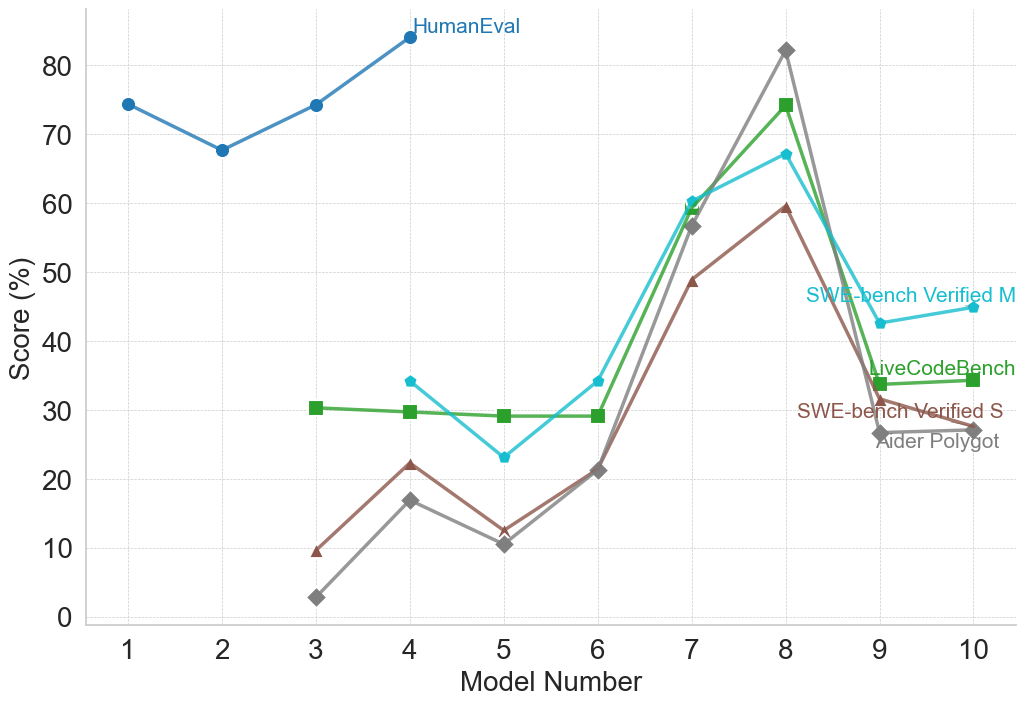}
        \caption{Programming and Coding}
        \label{fig:resK}
    \end{subfigure}

    \vspace{0.5em}
    \begin{subfigure}{0.45\textwidth}
        \centering
        \includegraphics[width=\linewidth]{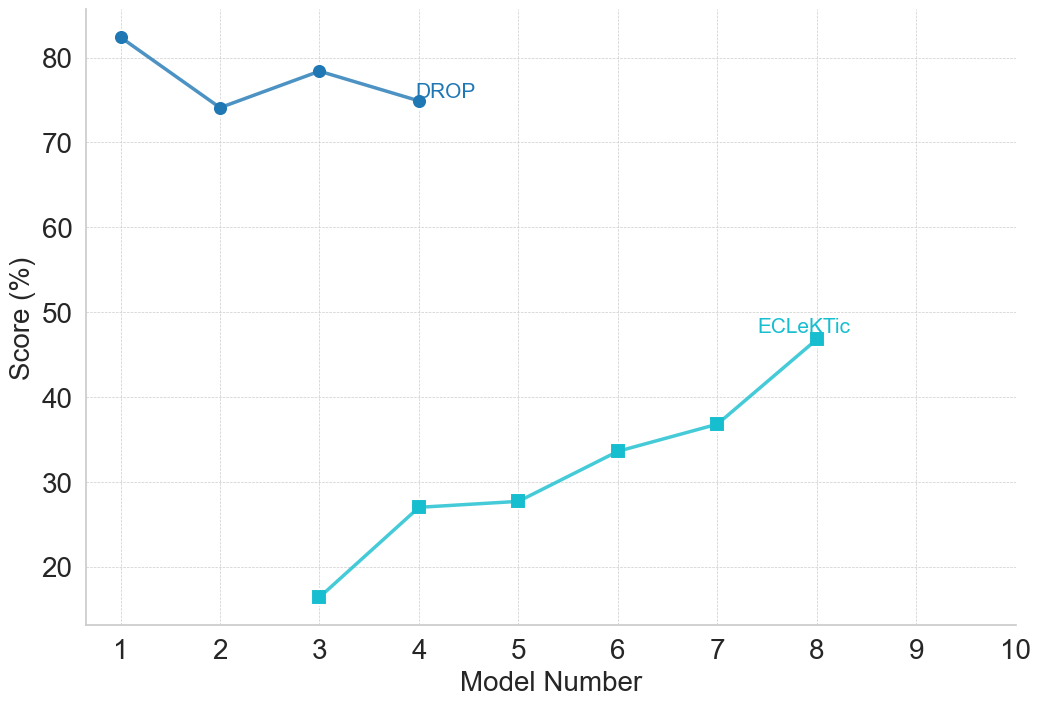}
        \caption{Reading Comprehension and QA}
        \label{fig:resL}
    \end{subfigure}
    \hfill
    \begin{subfigure}{0.45\textwidth}
        \centering
        \includegraphics[width=\linewidth]{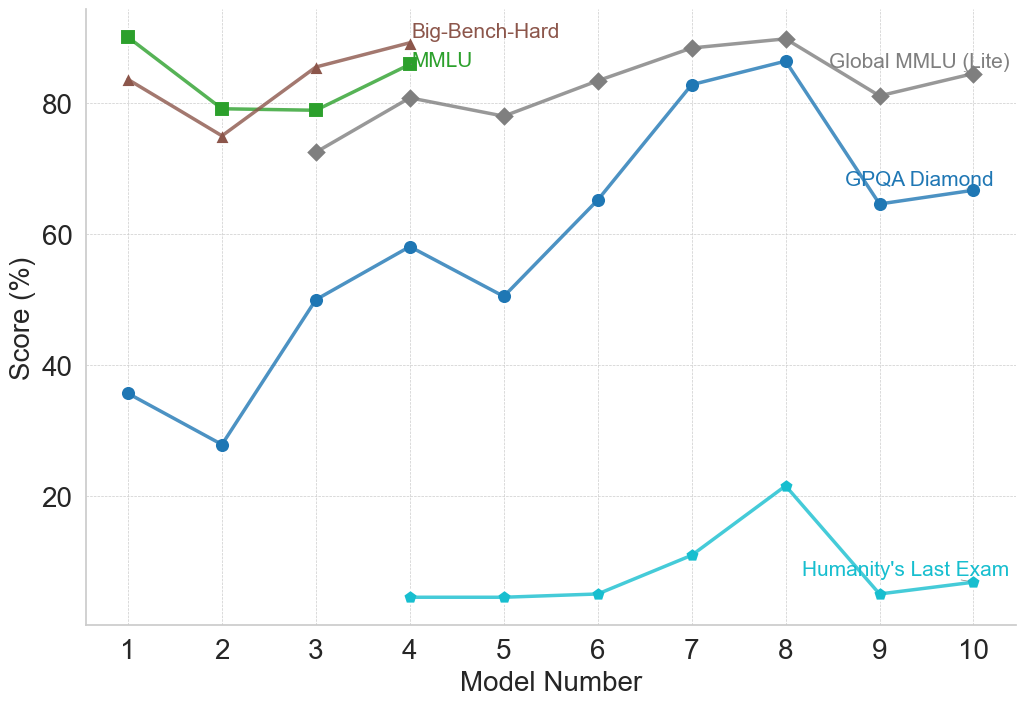}
        \caption{Reasoning with General Knowledge}
        \label{fig:resM}
    \end{subfigure}

    \vspace{0.5em}
    \begin{subfigure}{0.45\textwidth}
        \centering
        \includegraphics[width=\linewidth]{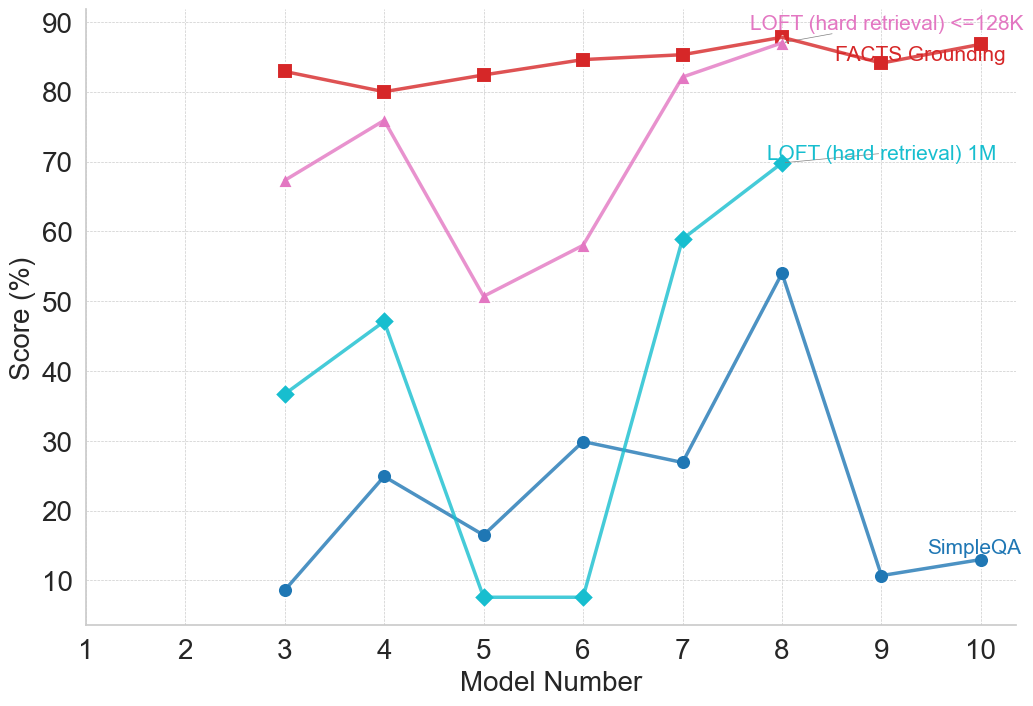}
        \caption{LLM Benchmarks}
        \label{fig:resNKB}
    \end{subfigure}

\caption{
Performance of the Gemini family on reasoning benchmarks by category. 
Model numbers and corresponding names are as follows: 
1 -- Gemini Ultra; 
2 -- Gemini Pro; 
3 -- Gemini 1.5 Flash; 
4 -- Gemini 1.5 Pro; 
5 -- Gemini 2.0 Flash-Lite; 
6 -- Gemini 2.0 Flash; 
7 -- Gemini 2.5 Flash; 
8 -- Gemini 2.5 Pro; 
9 -- Gemini 2.5 Flash Lite (no thinking); 
10 -- Gemini 2.5 Flash Lite (thinking).
}

    \label{fig:gemini-grid}
\end{figure}

\begin{figure}[htp]
    \centering
    
    \begin{subfigure}{0.45\textwidth}
        \centering
        \includegraphics[width=\linewidth]{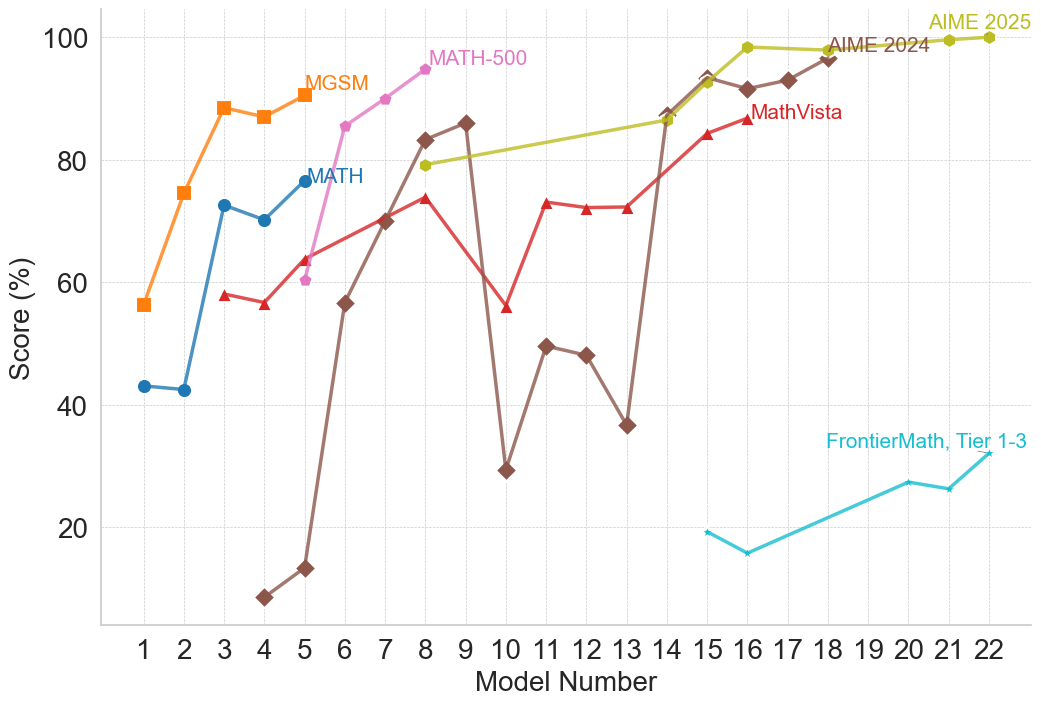}
        \caption{Mathematical Reasoning}
        \label{fig:resNK}
    \end{subfigure}
    \hfill
    \begin{subfigure}{0.45\textwidth}
        \centering
        \includegraphics[width=\linewidth]{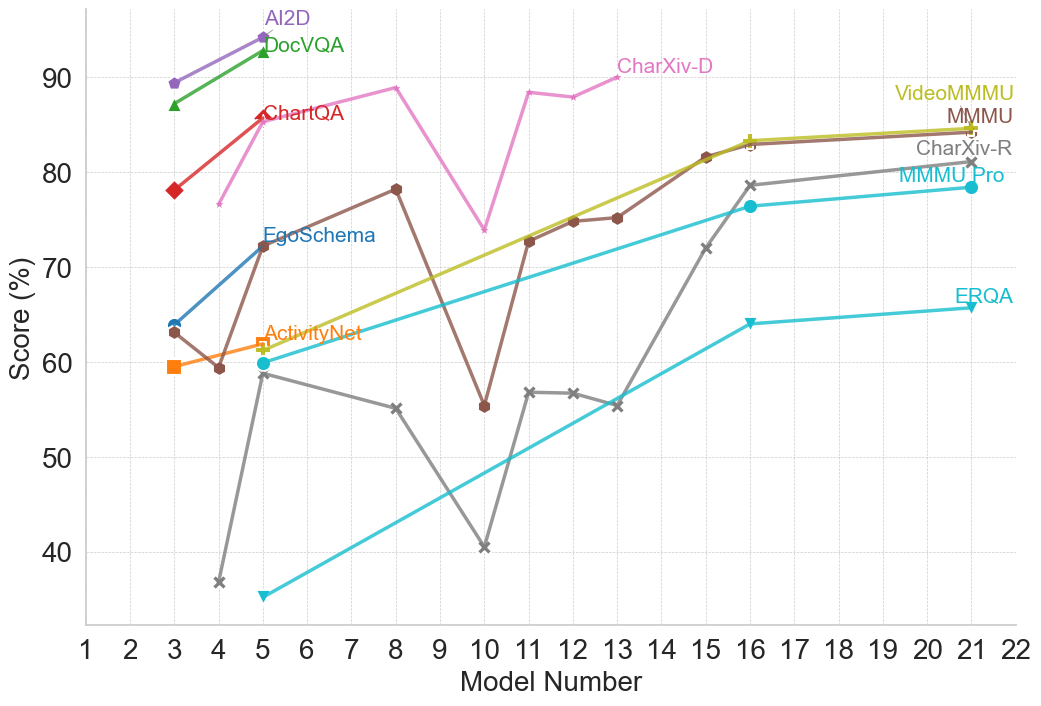}
        \caption{Multimodal Reasoning}
        \label{fig:resO}
    \end{subfigure}

    \vspace{0.5em}
    \begin{subfigure}{0.45\textwidth}
        \centering
        \includegraphics[width=\linewidth]{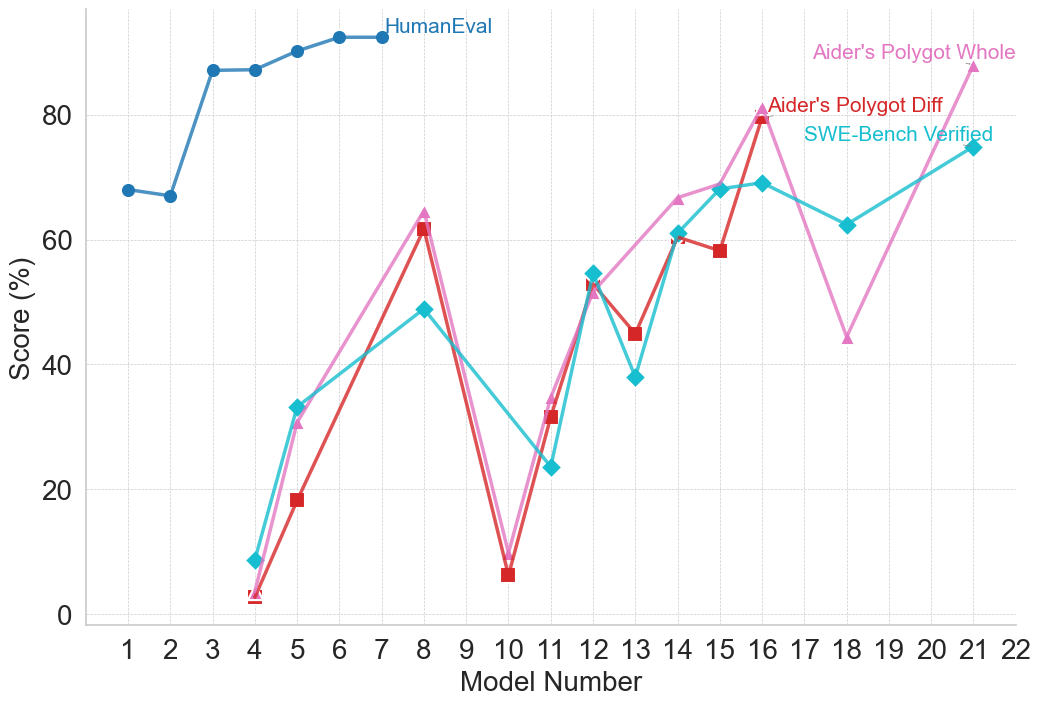}
        \caption{Programming and Coding}
        \label{fig:resP}
    \end{subfigure}
    \hfill
    \begin{subfigure}{0.45\textwidth}
        \centering
        \includegraphics[width=\linewidth]{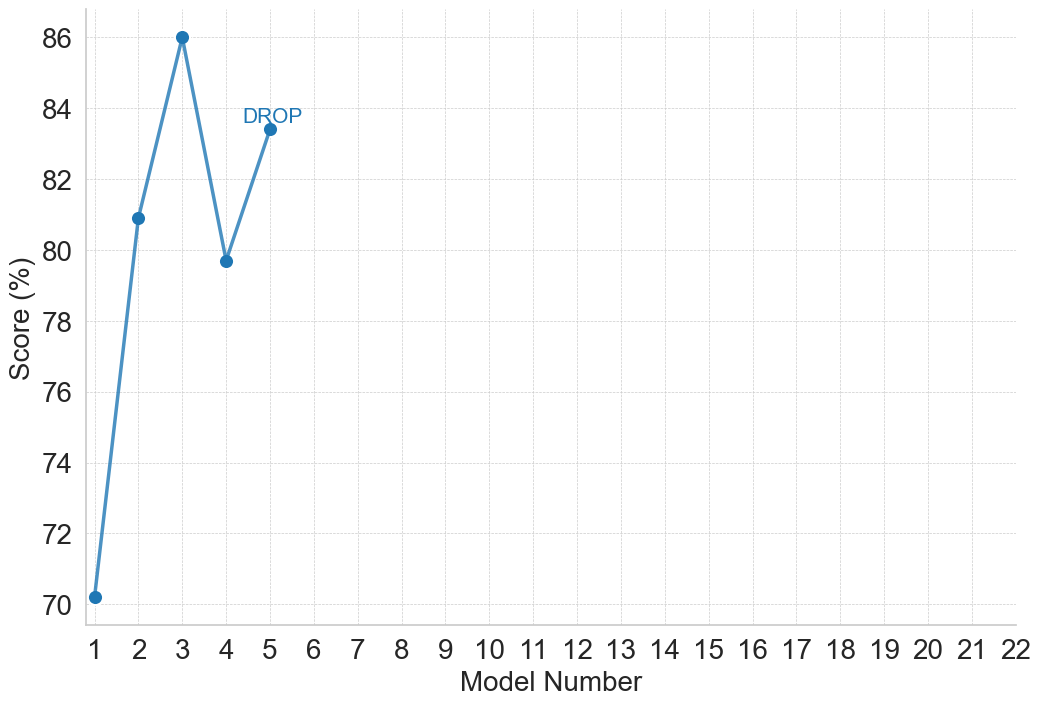}
        \caption{Reading Comprehension and QA}
        \label{fig:resR}
    \end{subfigure}

    \vspace{0.5em}
    \begin{subfigure}{0.45\textwidth}
        \centering
        \includegraphics[width=\linewidth]{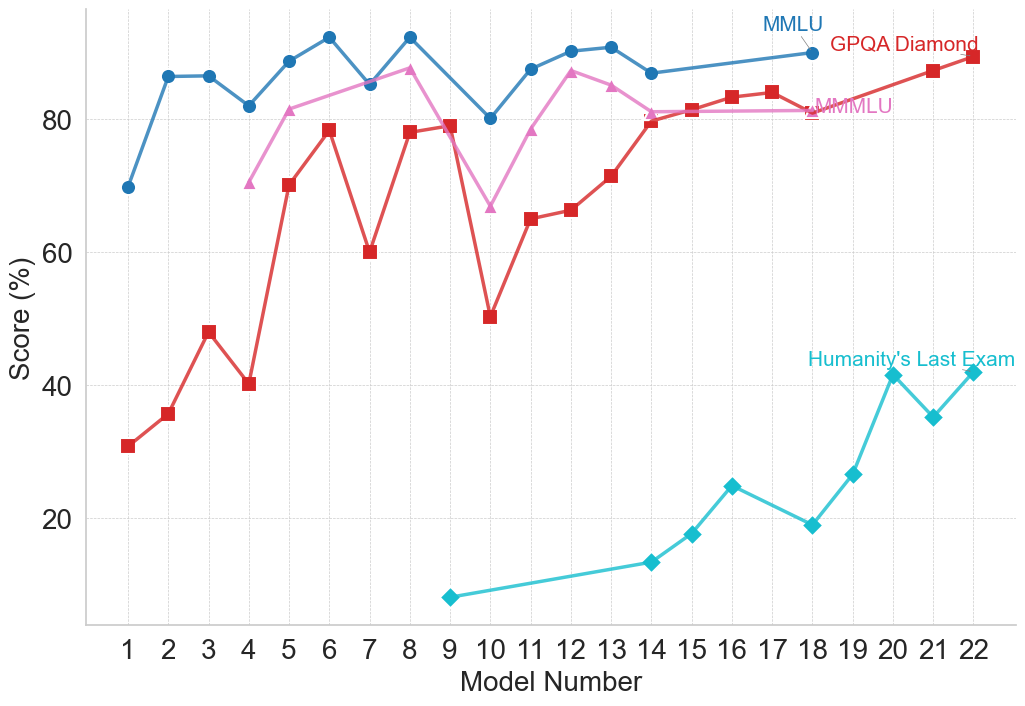}
        \caption{Reasoning with General Knowledge}
        \label{fig:resS}
    \end{subfigure}

\caption{
Performance of the GPT family on general reasoning benchmarks. 
Model numbers and corresponding names are as follows: 
1 -- GPT-3.5; 
2 -- GPT-4; 
3 -- GPT-4 Turbo; 
4 -- GPT-4o mini; 
5 -- GPT-4o; 
6 -- o1-preview; 
7 -- o1-mini; 
8 -- o1; 
9 -- o1-pro; 
10 -- GPT-4.1 nano; 
11 -- GPT-4.1 mini; 
12 -- GPT-4.1; 
13 -- GPT-4.5; 
14 -- o3-mini; 
15 -- o4-mini; 
16 -- o3; 
17 -- o3-pro; 
18 -- gpt-oss-120b; 
19 -- GPT-5 with Deep Research; 
20 -- ChatGPT Agent; 
21 -- GPT-5; 
22 -- GPT-5 Pro.
}

    \label{fig:gpt-nonllm}
\end{figure}

\begin{figure}[htp]
    \centering

    \begin{subfigure}{0.45\textwidth}
        \centering
        \includegraphics[width=\linewidth]{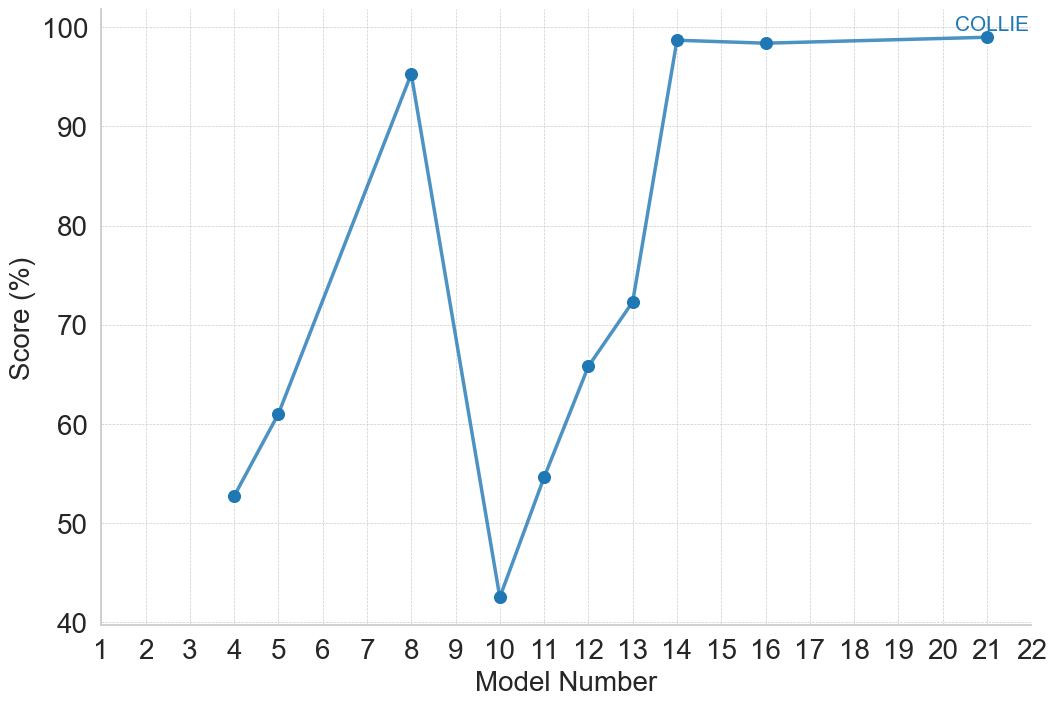}
        \caption{Constrained Text Generation}
        \label{fig:resMG}
    \end{subfigure}
    \hfill
    \begin{subfigure}{0.45\textwidth}
        \centering
        \includegraphics[width=\linewidth]{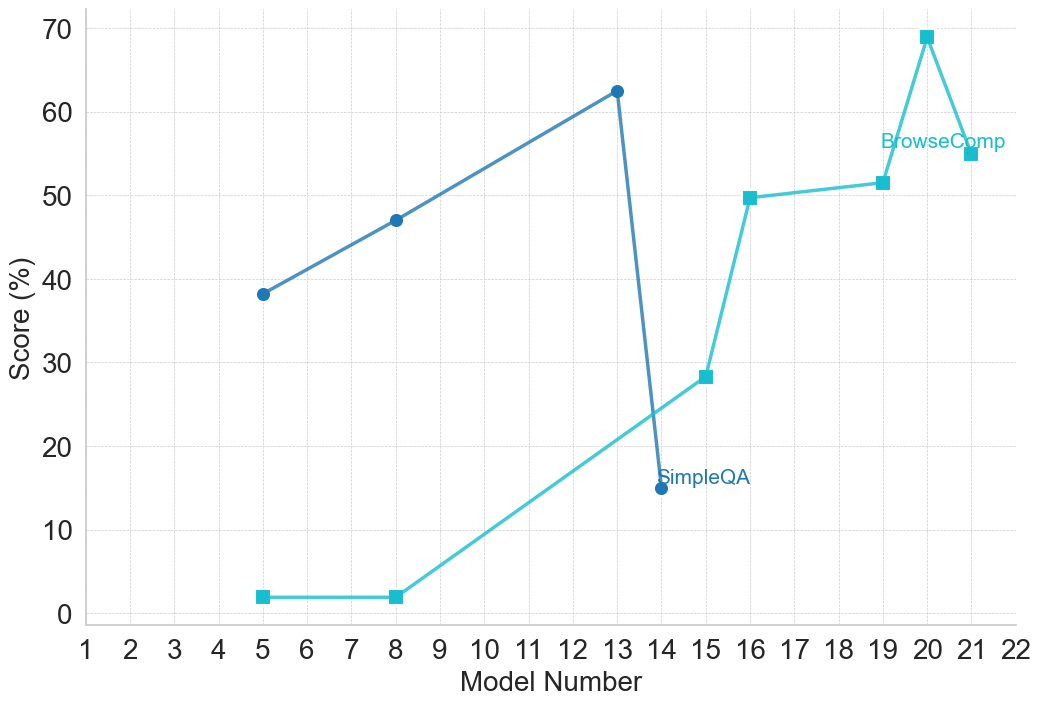}
        \caption{Factuality}
        \label{fig:resT}
    \end{subfigure}

    \vspace{0.5em}
    \begin{subfigure}{0.45\textwidth}
        \centering
        \includegraphics[width=\linewidth]{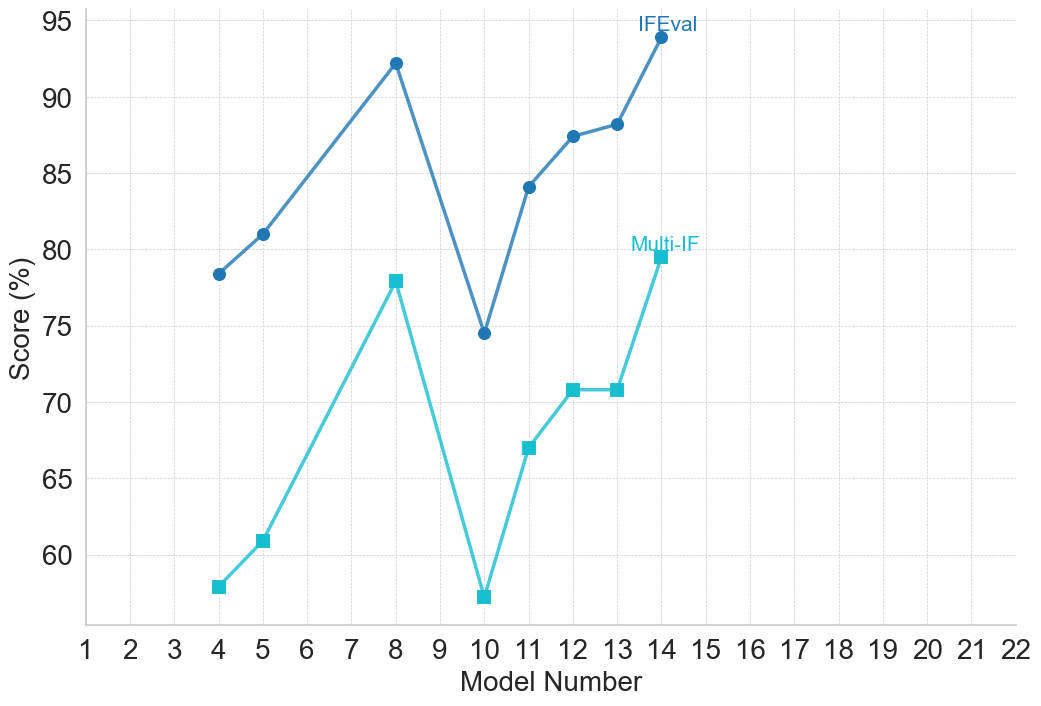}
        \caption{Instruction Following}
        \label{fig:resU}
    \end{subfigure}
    \hfill
    \begin{subfigure}{0.45\textwidth}
        \centering
        \includegraphics[width=\linewidth]{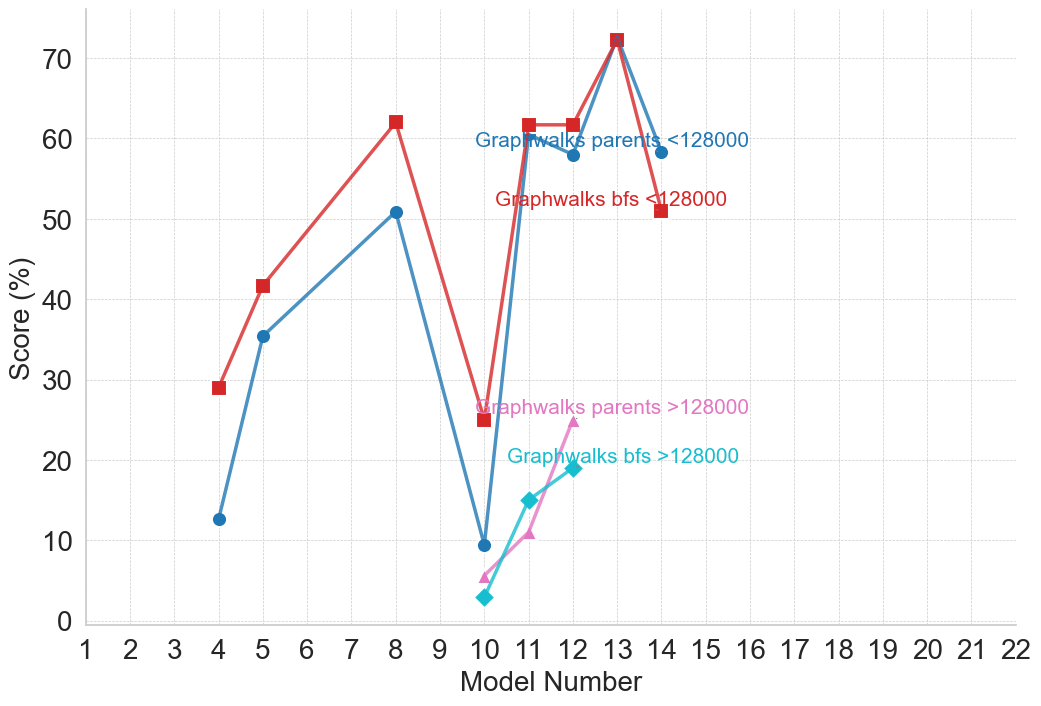}
        \caption{Long Context}
        \label{fig:resV}
    \end{subfigure}

    \vspace{0.5em}
    \begin{subfigure}{0.45\textwidth}
        \centering
        \includegraphics[width=\linewidth]{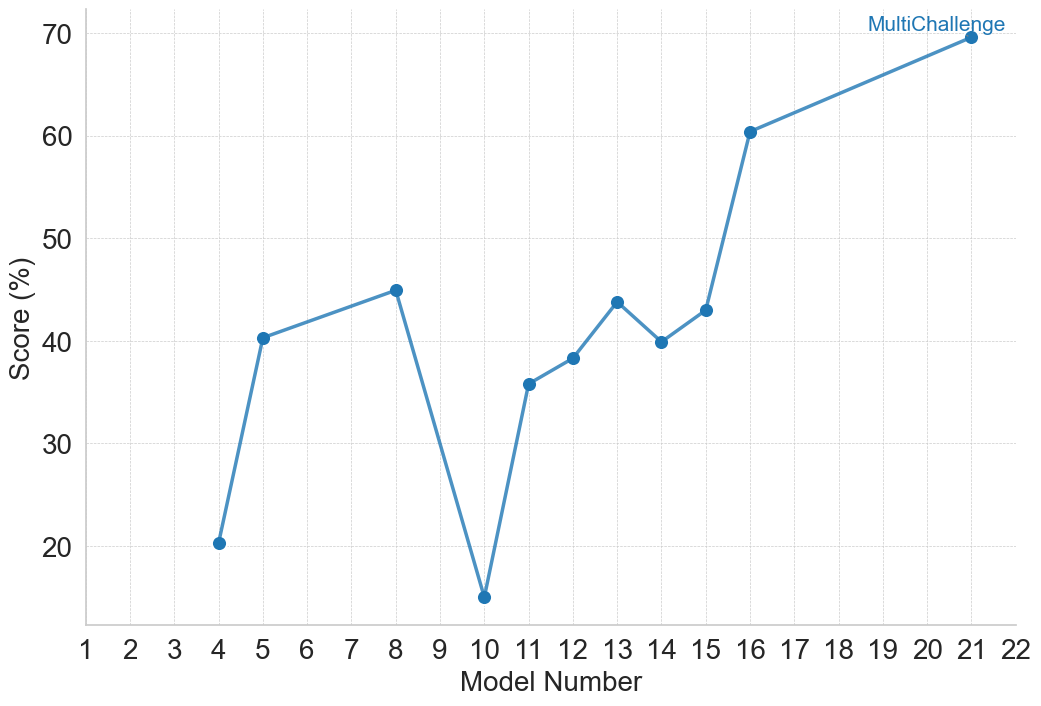}
        \caption{Multi-turn Conversation}
        \label{fig:resY}
    \end{subfigure}
    \hfill
    \begin{subfigure}{0.45\textwidth}
        \centering
        \includegraphics[width=\linewidth]{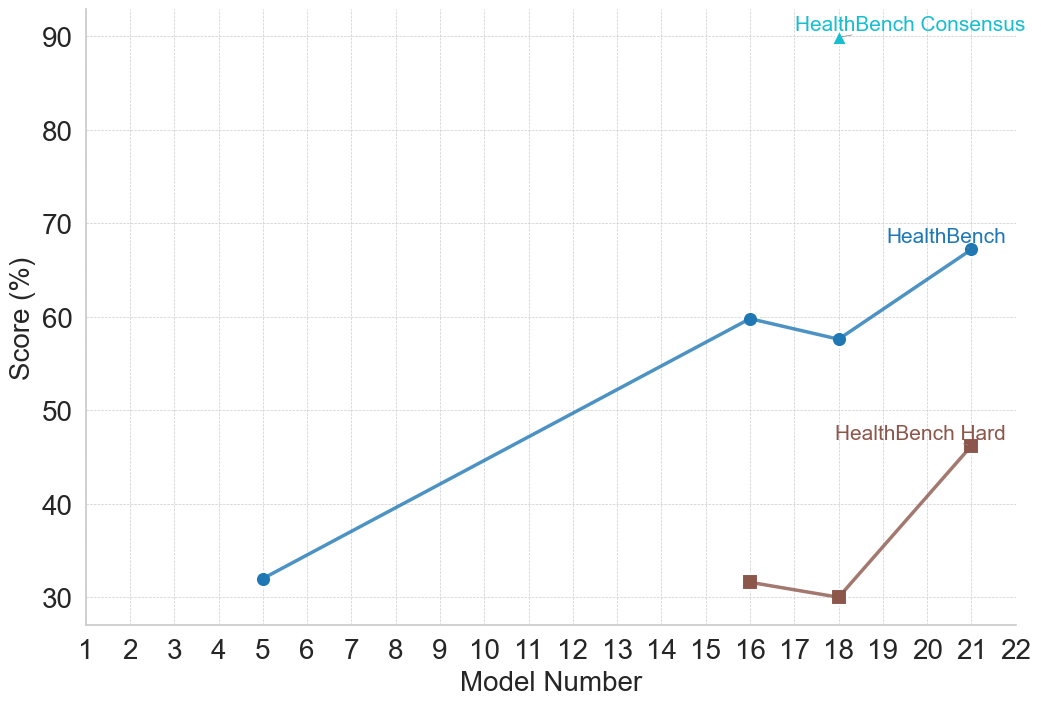}
        \caption{Safety}
        \label{fig:resZ}
    \end{subfigure}

    \vspace{0.5em}
    \begin{subfigure}{0.45\textwidth}
        \centering
        \includegraphics[width=\linewidth]{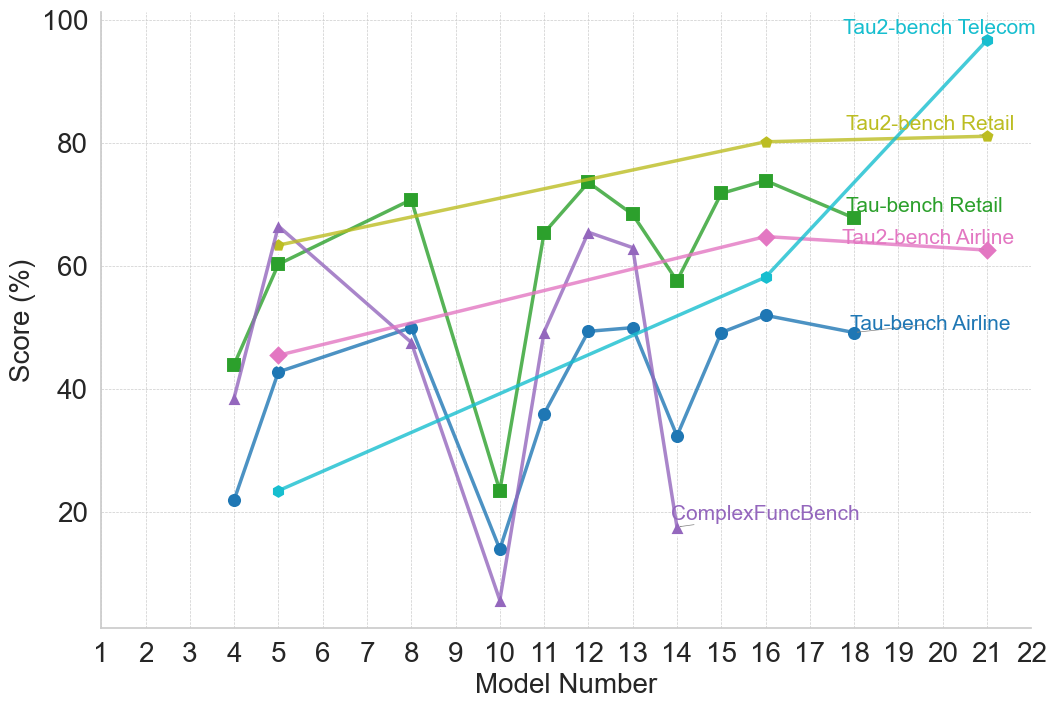}
        \caption{Tool Use}
        \label{fig:resQ}
    \end{subfigure}

\caption{
Performance of the GPT family on LLM-specific benchmarks. 
Model numbers and corresponding names are as follows: 
1 -- GPT-3.5; 
2 -- GPT-4; 
3 -- GPT-4 Turbo; 
4 -- GPT-4o mini; 
5 -- GPT-4o; 
6 -- o1-preview; 
7 -- o1-mini; 
8 -- o1; 
9 -- o1-pro; 
10 -- GPT-4.1 nano; 
11 -- GPT-4.1 mini; 
12 -- GPT-4.1; 
13 -- GPT-4.5; 
14 -- o3-mini; 
15 -- o4-mini; 
16 -- o3; 
17 -- o3-pro; 
18 -- gpt-oss-120b; 
19 -- GPT-5 with Deep Research; 
20 -- ChatGPT Agent; 
21 -- GPT-5; 
22 -- GPT-5 Pro.
}

    \label{fig:gpt-llm}
\end{figure}


\end{document}